# Design of the Anatomically Correct, Biomechatronic Hand

Benedek József Tasi, Miklós Koller, and György Cserey

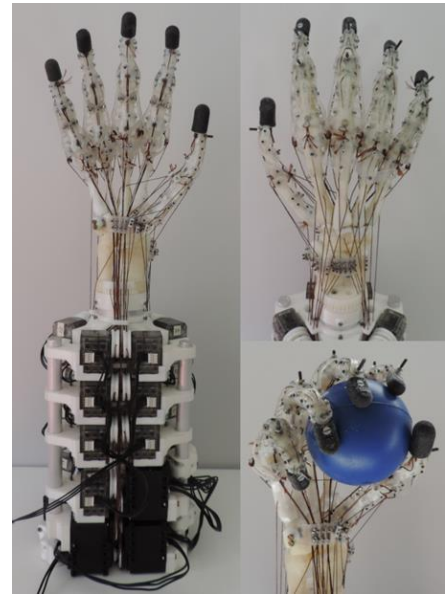

Fig. 1. The ACB-Hand with its forearm-based actuation system

*Abstract*—Following the recent advances in the field of robotic hand development, it can be inferred that the greatest challenge lies in restoring not just general functionality, but the delicate dexterity of the human hand. Results from studying its movement kinematics and dynamics indicate that the intricate anatomical features and structural details are essential for producing such versatility. To address this, we present the Anatomically Correct, Biomechatronic (ACB) Hand, by explaining the purpose and implementation of every functionally significant bone, ligament, finger-actuating intrinsic and extrinsic muscle, tendinous network and pulley system. We biomechanically validate our results using Landsmeer's models, and demonstrate the hand's performance by examining the execution of certain synergistic movements, analyzing fingertip trajectories, performing the Kapandji test, and realizing the GRASP taxonomy. This study aims to promote how anatomically accurate design considerations can assist in consequentially attaining more human-like functionality, while still allowing for robust structural elements towards advancing research in hand biomechanics, prosthetics and teleoperation.

## I. Introduction

MOST existing anthropomorphic robotic hands are designed to restore the primary abilities of the human hand, like simple grasping and producing basic gestures. This is usually achieved by following the conventional approach of employing mechanized joints and an actuation system with significantly altered structure and often reduced biomechanical complexity compared to the biological joints and tendinous network. Some of the most advanced examples for this are the Shadow Dexterous Hand of the Shadow Robot Company [1], and the hand of the DLR Hand-Arm System, developed by M. Grebenstein *et al.* [2]. The degrees of freedom (DoF) and overall capabilities of these models are useful for approximating the behavior of the human hand, and perform tasks requiring a promising level of dexterity, but they simplify the internal anatomical structure and biomechanical relations, focusing on their apparent functional properties. This would reduce their efficiency within the boundaries of prosthetics or manipulators controlled directly by a human through teleoperation, as complex transfer functions and heavy computation would be required to adequately translate human intentions and individual muscle activities into motor commands for a differently actuated device, while preserving the biologically defined active and passive dynamical relations. This would make direct, dexterous control difficult, as the operator inevitably loses the natural, "organic" feeling of motion, and has to adapt to a vastly different system.

We believe that it is possible to improve upon these concepts by realizing the elaborate anatomical structure of the hand, which is essential for human-like dexterity. However, due to the high complexity and natural variance of these features, and the fact that there is still no real consensus on the exact biomechanical utility of several details, there is only few research attempting to fully implement them in robotic models. Additionally, while the selective inclusion of the components with the most prominent functionality has proven to be sufficient for producing some grasps and gestures in a dexterous and somewhat biomimetic way, the detailed functional anatomical properties of the hand [3] – resulting from the abundance of intrinsic and extrinsic muscles and the varying geometrical features of the musculoskeletal and ligamentous

This research has been partially supported by the European Union, co-financed by the European Social Fund (EFOP-3.6.3-VEKOP-16-2017-00002), Jedlik Innovation Kft., and the Info-bionics Association.
Authors are with the Information Technology and Bionics Faculty of Pázmány Péter Catholic University, Budapest (Hungary) (e-mail: tasi.benedek.jozsef@hallgato.ppke.hu, koller.miklos@itk.ppke.hu, cserey.gyorgy@itk.ppke.hu).

system – are highly critical for producing precise fine-motoric movements, like handwriting or playing musical instruments, and thus each should be accounted for.

One of the first, most widely used hand model following in these footsteps with a more anatomically inspired approach, the ACT Hand [4] incorporates a crocheted realization of the tendinous system for the first three fingers, which mimics the kinematical behavior of biological structure, and thus provides a compliant platform for examining hand movements in detail. It consists of internally placed, fixed-axis hinge and gimbal joints, which allow for the examination of muscle activity effects in an ideally constrained joint system, removing the dynamical issues with biological ligaments.

Based on this project, Z. Xu and E. Todorov have developed the Highly Biomimetic Anthropomorphic Hand [5], which implements the extensor network and the flexor pulleys, along with crocheted ligaments in an ergonomic design with dexterous capabilities. This model also successfully demonstrates the dominance of the most significant muscles of the hand, by employing certain simplifications towards eliminating redundancy like the passivation of the interosseous, exclusion of the FDS and lumbricals, partial exclusion and functional merging of the intrinsic and extrinsic thumb muscles, and the modification of their tendinous network. The joints are actuated by strings connecting the actuators directly to the phalanges, while the laser-cut extensor network serves as a ways to harness and distribute torques to different areas of the finger, thus creating an under-actuated system, capable of efficient grasping and dexterous object manipulation with remarkably human-like performance.

The Human-like Robotic Hand, presented by A.A.M. Faudzi *et al.* [6] for clinical educational purposes, is a significant step in the direction of total anatomical accuracy. In this model, the complete joint capsule is preserved by silicone strips glued to the bones, and all notable intrinsic and extrinsic muscles are implemented using thin pneumatic McKibben artificial muscles. This helps in the realization of not only the biomechanical properties and capabilities of the hand, but the volume and orientation of its musculoskeletal system as well, since it allows for anatomically accurate muscle placement. This approach is extremely valuable in examining the contribution of each muscle to different hand and finger movements, and it efficiently preserves the volume and passive behavior of the biological hand as well.

Our goal is based on merging the achievements of these two designs: the purpose of this study is to thoroughly understand and model the intricate structure of the human hand, which sources its extreme dexterity and precision, and develop a biomechatronic model that includes the interpretation of every distinguishable, functionally contributing anatomical detail like articular surface geometry, an accurate representation of the tendinous and ligamentous system alongside every intrinsic and extrinsic muscle, while providing a robust template with an optimized fabrication process towards the development of dexterous manipulators and prosthetics with more direct control and human-like behavior. We believe that understanding the exact biomechanical background of finger movements, down to

TABLE I
COMPARISON OF EXISTING ANATOMICAL HAND MODELS

| Robotic hand | Highly Biomimetic, Anthropomorphic Hand [5] | Human-like Robotic Hand [6] | Anatomically Correct, Biomechatronic Hand (developed) |
|---|---|---|---|
| Bones | 3D printed, ABS | Molded, resin | 3D printed (PolyJet), resin |
| Actuation | 10 servo motors | Pneumatic McKibben actuators | 30 servo motors |
| Extrinsic muscles | FDP, EDC included, FDS, EI, EDM omitted. FPL included, APL, EPB, EPL merged | All included, except EI and EDM | All included, except EI and EDM |
| Interossei | Passivated and fixed to preserve poise and regulate add/abd through elastic silicone | All of them included in their biological location | All of them included, external actuation through bone tunnels |
| Lumbricals | Not included | All included in their biological location | All included, focus on synergy. External actuation through bone tunnels, FDP dependency realized digitally |
| Ligaments | Crocheted ligaments, PCL-ACL functionally merged, volar plate included as check-rein bands | All included (PCL, ACL, volar plate), silicone ligaments glued to the bones | All included (PCL, ACL, volar plate), silicone ligaments screwed to the bones |
| Flexor pulleys | Grommeted, laser-cut elastic silicone | Polyethylene tubes | PTFE tubes fixed to elastic silicone pulleys |
| Extensor network | Laser-cut silicone extensor network and ligaments, assistive (torque-distribution) role to the primarily string-based actuation | Dyneema tendinous network, ligaments included | Silicone extensor network laser-cut from a single piece, deep slip and auxiliary ligaments included, silicone network handles actuation, strings only connect proximal endpoints and servos |
| Hypothenar muscles | Passivated and fixed, under-actuated palm bending | All of them included, thin muscles | Slaved together for producing their average effects and preserving direction |
| Thumb/ Thenar muscles | Three muscles, flexor, adductor, extensor, extensor network similar to the long fingers | Each head included, bundled extensor expansion | All included separately, with multiple-heads, laced through bone tunnels. Laser-cut unique thumb extensor expansion. |
| Sesamoids | Average sesamoid muscle insertion in the bone location | Muscles connected to sesamoid locations | Sesamoid bones realized as toruses, tethering muscles and silicone ligaments |

Table I contains the detailed comparison of the examined anatomically inspired hands and our own designs, highlighting the included features and the most important differences.

individual muscle activity is essential for the faithful reproduction of elaborate fine-motoric performance. To this end, thorough comparison has been made regarding the two previously described hand models, to characterize the required details and advantages for our prototype. This is summarized in Table I. The need is also recognized for averaging the wide range of different anatomical variations of the features occurring in the human hand [7], [8], [9], [10], and for attempting to preserve the characteristics of the connected structure of the functionally independent compartments. An engineered model will always be constructed from independently fabricated parts, whereas in biology, everything is coadunate to a certain level, which must be properly translated into the compartmental robotic design.

As the anatomical diversity of the human hand implies that each component has a generally well-definable set of functional aspects, which are more important than the extreme spatial variations or specific kinematical accuracy, we settle on a more empirical approach: Instead of designing the model primarily with the desired skillset and mechanical capabilities in mind, the direction is to accurately mimic the anatomical structure, engineer an efficient mechatronic actuation system to substitute the biological muscles, then evaluate how the complete mechanism responds to individual, sequential or synergistic-antagonistic actuation, and refine our design accordingly to eventually approach human-like behavior. We believe that for synergistically complex and precise movements, like handwriting or playing musical instruments, the importance of the synergistic-antagonistic actions of all finger-actuating muscles must be emphasized [3], and if the goal is the complete restoration of hand functionality and the investigation of dexterity, the effects of each should be taken into consideration during the design process.

In addition to our objectives for application in the field of dexterous manipulators, our final, anatomically focused and biomechanically validated results could lay the foundations for advanced prosthetic and biomedical research, most notably by serving as a tool in the functional characterization of hand performance and injuries, medical education and surgical preparations, the development of advanced neural interfaces to identify detailed muscle activity or process motor signals for prosthesis control, and to study the muscle activity patterns of the hand during different precision movements towards their reproduction on an algorithmic level.

The structure of this paper is the following: We present the characteristics of our life-sized prototype, shown in Fig. 1, and explain the fabrication of our artificial interpretation of the relevant anatomical details and the resulting functionality. We also introduce an actuation system, which substitutes every important finger-actuating muscle with one of three different types of Dynamixel servos. Then we perform a detailed analysis of synergistic muscle activity, and validate our results using the biomechanical models of Landsmeer, and by performing the Kapandji test. Finally, we demonstrate the performance of our model through the examination of fingertip trajectories and several grasping examples.

## II. ANATOMICAL DESIGN

In this section, we detail the anatomical features implemented in our prototype. We explain the biomechanical significance of each component, and show how our implementation corresponds to the physiology of the biological hand. We aim to present a design which honestly translates the anatomical features to a biomechatronic system, without hindering the dexterity ensured by the synergistic activity and coadunation of the biological components.

### A. Modeling the bones

The human hand contains 27 bones: one metacarpal (MC) and three phalanges (proximal – PP, middle – MP and distal – DP) for each finger, with the notable exception of the thumb, which is one phalanx shorter. The wrist is formed by 8 small carpal bones, which connect the fingers to the ulna and radius. Between each consecutive bone pair, one of two possible joint types is formed, distinguished by their range of motion, which is primarily constrained by the articular surfaces of the contacting bones. The metacarpophalangeal (MCP) joints generally have two degrees of freedom (DoF), allowing for flexion, extension, adduction and abduction, while the interphalangeal (proximal – PIP, distal – DIP, or simply IP for the thumb's only joint of this type) joints have only one.

Although some of the more advanced hand models correctly implement the aforementioned joint DoF, most of them treat the fingers as parallel running links of differently sized phalanges, connected by hinge joints [11], [12], [13]. While providing a stable and highly efficient base for various finger-actuating mechanisms, this solution strictly limits the movement range of the joints outside their active planes. It is important to point out that the ability to firmly wrap our fingers around objects depends on enabling our fingers to conform around a multitude of shapes [3]. In reality, the axis of flexion-extension at each joint is increasingly oblique from the index to the little finger, from proximal to distal direction, resulting from the slight asymmetry and rotation of the bone heads around their shaft axes, which helps the hand form an oblique, tunnel-like geometry bounded by the palm and the fingers during heavier grips. The IP joint of the thumb is not only oblique, but its center of revolution cannot be truly defined as an axis, but rather as a cone: the anterior protrusions of the bone head cause a slight automatic medial axial rotation during flexion, promoting thumb opposition. Additionally, due to the asymmetric anterior swellings on the MC heads, abduction and adduction also occurs around an oblique, conical axis at the MCP joints, resulting in the possibility of a small degree of active or passive axial rotation.

To sufficiently preserve the constraints set by the articular surfaces, and thus the unique structure of each joint type, the bones were 3D printed from hard VeroWhite resin using PolyJet technology to attain extreme surface smoothness, using open-source models from a database generated from the averaged results of multiple MRI scans [14].

As wrist movements are out of the scope of this study, the third joint type, the carpometacarpal (CMC) joints are immobilized: the carpal and metacarpal bones are fused together, with the exception of the trapeziometacarpal (TMC) joint of the thumb, which plays a critical role in thumb opposition (as explained later). Preservation of the exact geometry of the wrist, however, is of great importance, along with the heads of the ulna and the radius, as each bone contributes to the formation of the tendon pathways and muscle arrangement, and consequently, their shape affects force directions and efficient torques.

*B. Joints and ligamentous structure*

The contacting bones are held together by the joint capsule, which consists of two fibrous bundle groups with different functionalities. These are shown on Fig. 2/(a). The proper collateral ligaments (PCL) are the main bundles tethering the adjacent bone heads and bases together. Their position contributes to constraining the flexion-extension axis, and their level of tension is responsible for further limiting the joint's range of motion. Each phalanx base is extended by a fibro-cartilaginous volar plate, connected to it by a flexible recess. The plate is suspended from the joint's bone head by the accessory collateral ligaments (ACL). The fibers of the PCL and ACL arise from the same area, but the ACL tethers to the phalanx base slightly anteriorly. During flexion, this results in the volar plate being pressed against the anterior surface of the bone head under the pull of the ACL, stabilizing the joint.

To enable adduction and abduction at the MCP joint, its PCL are lax in extension, as opposed to their constant stiffness at the IP joints. Their origin point, however, is located slightly posterior to the primary axis of the joint, which results in the PCL stiffening up during flexion, when the PP base slides over the palmar protrusions of the MC head, preventing side-to-side movement in a flexed state.

To preserve these geometric limit sets, the ligament bundles were cut from 1 mm thick silicone sheet of ShoreA60 hardness, chosen for its adequate flexibility and limited elasticity. The PCL and ACL were cut from a single piece on each side to reduce lateral protrusion upon being screwed to the bones in their anatomical insertion points, as shown in Fig. 2/(b). The volar plate is realized as a wider, harder (ShoreA80) silicone piece, the distal end of which is embedded into the phalanx bases to prevent the obstruction of the flexor tendon pathways. At the MCP joint, it is fabricated independently, and connected to the ACL via an M1 screw and nut combination. This permits their rotation around each other, which accounts for the loss of individual fibers tightening separately during the tensioning of the ligaments, caused by the homogeneity of the silicone. This results in the volar plate bending in the correct direction under the pull of the ACL, when it is pressed against the anterior bone surface during flexion.

The role of elasticity and passive joint resistance is often overlooked despite their importance in muscle activity based movement coordination. Our joint design aims to enhance our model's capability for naturalistic motion, recognizing the need for mimicking the behavior of the human hand under passive strain just as well as during active movements.

*C. Muscles and the tendinous system of the long fingers*

The movements of the fingers are actuated by two spatially separate muscle groups: the extrinsic muscles, the bellies of which are located in the forearm, and the intrinsic muscles, with their complete volume laying in the palmar area. The extrinsic muscles are generally responsible for effectuating larger movements. They can be further categorized into two functionally different groups: the ventrally located long flexors, which bend the fingers, and the dorsal long extensors, which are used for straightening them.

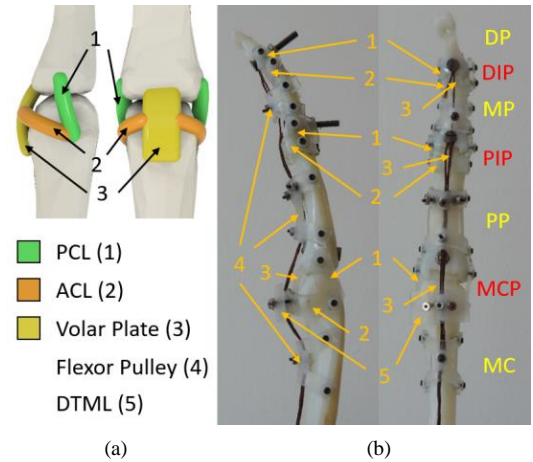

Fig. 2. The ligamentous system of the fingers. (a) Color-coded CAD model of the ligaments of the MCP joint, lateral (left) and dorsal (right) view. (b) The 3D printed artificial finger model with silicone joint ligaments, and silicone flexor pulleys reinforced with PTFE tubes, lateral (left) and dorsal (right) view.

There are two significant extrinsic flexors of the long fingers: the flexor digitorum profundus (FDP, deep layer) and sublimis (FDS, superficial layer). Their tendons separate before passing through the carpal tunnel below the flexor retinaculum (FR), and continue coursing under the elastic pulley system on the anterior side of the metacarpal bone and the three phalanges. The FDP inserts into the base of the DP, and the FDS into the base of the PP after perforating the FDP to avoid lateral displacement. Thus, they primarily flex the DIP and PIP joints respectively, but they have a secondary flexor effect on all of their preceding joints, including the wrist. The elasticity of the pulleys allows for their bulging under increased strain, which enables us to strengthen our grip more efficiently by increasing the flexor moment arms at the joints. The pathways are further optimized by the continuation of the MC pulley within the deep transverse metacarpal ligament (DTML) on the anterior side of the volar plate, promoting MCP flexion.

The tendons of the long extensor muscle, the extensor digitorum communis (EDC) separate around the level of the wrist, course under the extensor retinaculum (ER), and insert into each finger at multiple points. Its primary function is the extension of the MCP joint via its deep insertion into the PP base, but it splits into medial and lateral bands distally, which insert into the MP and DP bases respectively). These enable it to partially extend the PIP and DIP joints as well, before the deep slip fully stiffens, but this can be antagonized by the long flexors' activity on the IP joints to coordinate independent MCP and IP flexion. The index and little fingers each have an additional extensor muscle that fuses with the EDC around the level of the MCP joint, the extensor indicis (EI) and the extensor digiti minimi (EDM) respectively.

Due to the partially shared muscle bellies (each of the three primary extrinsic muscles has a shared belly in the forearm, which separates slightly before their tendons are given off), full flexion-extension of an individual finger is usually difficult without the muscle at least slightly affecting the other, neighboring digits. The most prominent flexor is the FDP, and the primary role of the EDC in this relation is to limit the level of flexion by antagonizing the flexors at the MCP. The FDS can either strengthen the grip (especially at the MCP [15]), or apply a secondary flexing force to the fingers, by bending some of them when others are kept extended in certain gestures. Thus, as the structure of the extrinsic muscles is relatively simple, their actions and effects are broadly well-defined, they cause impactful movements in a wide range, and they also actuate the wrist to a significant degree, it becomes apparent that the intrinsic muscles are the ones primarily responsible for the remarkable dexterity and an important portion of the hand's gripping strength. The coordination of independent flexion and extension of our fingers is possible as a result of the synergistic and antagonistic activity of the intrinsic and extrinsic muscles, where the former constrain the general effects of the latter.

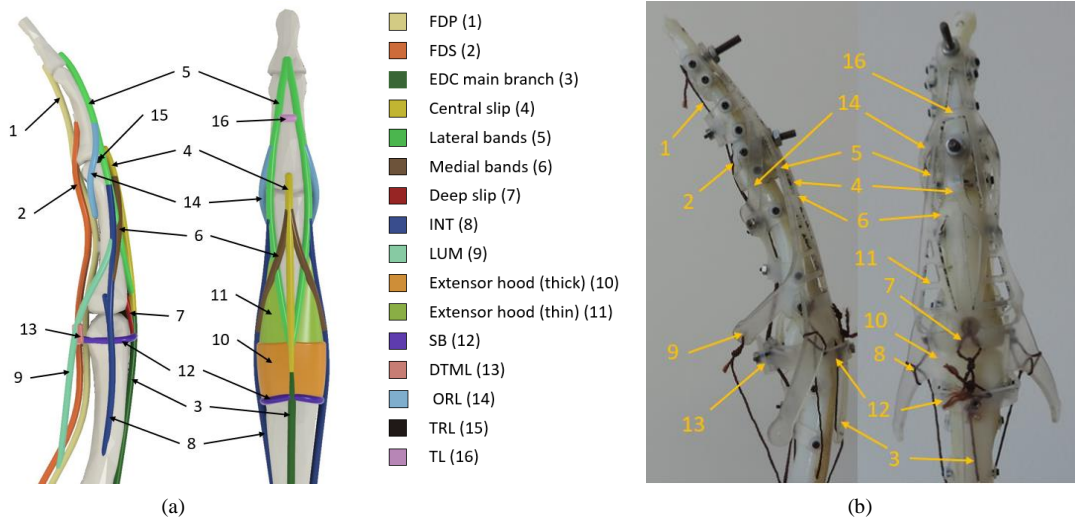

(a)  (b)

Fig. 3. The tendinous system of the long fingers. (a) Color-coded CAD schematics of our interpretation of the tendons and auxiliary ligaments of the index finger, without the flexor pulleys, from lateral (left, with the extensor hood hidden) and dorsal (right) view. Note how the medial and lateral bands overlap slightly, forming two different layers. (b) Our 3D printed finger with its laser-cut tendinous system, from lateral (left) and dorsal (right) view. Note that the overlapping lateral and medial band areas are averaged into a single merging point while preserving their pathway geometry to enable laser cutting from a single sheet. As a result of the natural resistance of the single-piece silicone network, and the presence of a screw head placed at the PIP joint, the sliding-constraining effects of the TRL were realized without the inclusion of a dedicated silicone band.

There are two, equally important subtypes of intrinsic muscles: the interossei (INT) and the lumbricals (LUM). The INT, which arise from the shafts of the metacarpals on both sides – besides being responsible for the adduction and abduction of the fingers via their first insertions into the sides of the PP base – have extending effects on the IP joints when the MCP is extended by the EDC, or flexion enhancing effects on the MCP joint when it is being flexed. In the latter case, the deep slip relaxes, so the EDC's activity can fully extend the IP joints [16], while at the same time, the INT can axially rotate (pronate or supinate) the finger at the MCP to a slight degree. To produce this complex functionality, the INT merge with the EDC tendon to form an elaborate, multi-layered tendinous structure that wraps around the posterior aspect of the finger: the extensor network [17], [18], [19]. This system, shown in Fig. 3/(a), is responsible for distributing the torques of the long extensors and the intrinsics to coordinate joint movements.

The LUM muscles originate from the palmar section of the FDP tendon, and insert into the lateral bands proximally from the extensor hood. They are weak flexors (flexor starters) of the MCP and extensors of the IP joints [20], but their efficiency does not depend on the degree of flexion at the MCP joint, due to their tendons coursing anteriorly to the DTML. The exact role of the LUM in specific grasping and dexterous tasks and is yet to be understood completely, and so they are often neglected, despite having an important part in the coordination of independent finger flexion-extension. Their synergy with the EDC allows for the extension of the IP joints of any individual finger, while maintaining the degree of flexion in every other digit and the MCP, due to them acting as active extensors as a consequence of their physiology. Instead of requiring the relaxation of the FDP to be efficient, the LUM are reinforced by its contraction, as they transfer the strain from its distal portion to enhance IP extension.

The system is further supplemented by five notable ligaments [21] that constrain the different bands to their workspaces, and contribute to the overall smoothness of motion. The sagittal bands (SB) prevent the EDC from slipping between the metacarpals during flexion, the triangular (TL) and transverse retinacular ligaments (TRL) constrain the sliding of the lateral bands, and the DTML separates and constrains the pathways of the interosseous and lumbrical muscles. The oblique retinacular ligament (ORL) assists the synergistically active muscles in the production of simultaneous IP joint rotation by automatically pulling on the distal portion of the lateral bands upon PIP extension.

The complexity of the extensor system shows that the fingers and the joints of a single finger are dependent on each other in a very specific way. It must be emphasized that the intrinsic muscles are not only synergizing with the extrinsics to coordinate fine-motoric movements, but they are simultaneously strongly contributing to independent finger activity as well, which is why the human hand has a certain, innate approach to grasping, and its biomechanical features cannot be truly substituted by simple mechanics that drive each joint independently from each other.

In our robotic hand, individual Dynamixel smart servo motors with built-in absolute encoders substitute the common extrinsic muscles separately for each finger, which allows us to set the level of coactivity between these muscle compartments digitally, enabling us to study the effects of certain anatomical variations, without limiting the design to one concrete realization. Our actuator system is primarily based on the XL-320 servo series, which poses a good balance between compactness, precision, resolution, speed and torque. Higher grade, MX-12W servos were used at the FDP of the index and middle finger, capitalizing on their ability of multiple controlled revolutions, with four times the resolution of an XL-320. For the sake of simplicity, the INT of the ring and middle finger were tethered to a single XL-320, as their side-to-side movements are usually negligible. Furthermore, the FDS of the ring and little fingers were slaved together, as their role is mostly restricted to grip reinforcement, and this muscle is more redundant, and often found missing in the human little finger.

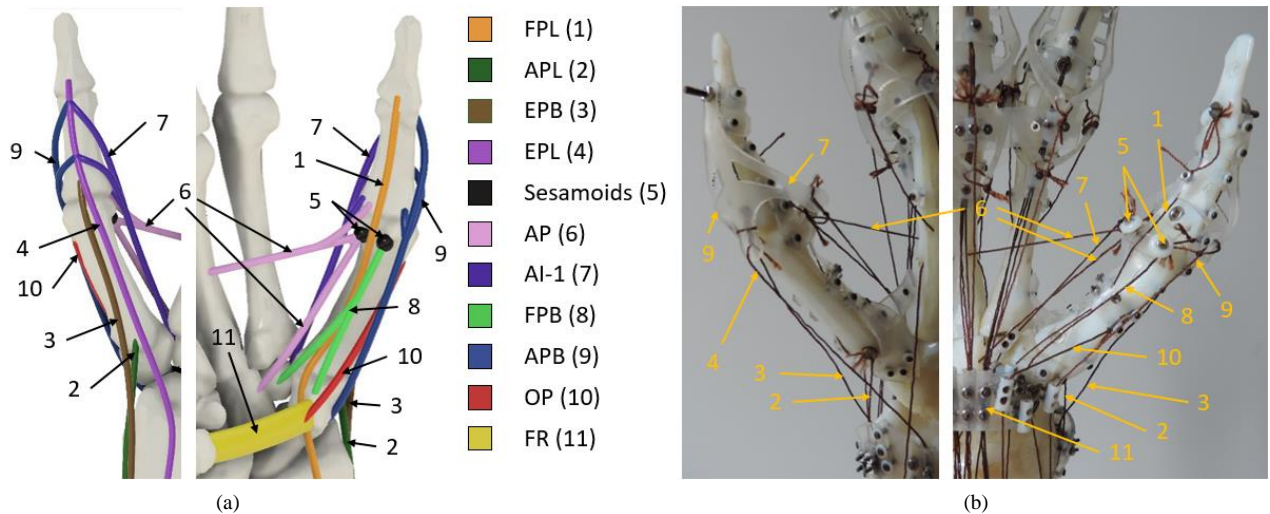

Fig. 4. Musculoskeletal system of the thumb. (a) Color-coded CAD schematics of the extrinsic and intrinsic thumb muscles, showing all their heads and insertions, dorsal (left) and ventral (right) view. (b) Our anatomically correct thumb design, showing the PTFE tendon guides on the bone surfaces and the flexor retinaculum, dorsal (left) and ventral (right) view.

The anatomically correct inclusion of the intrinsic muscles in the palm, however, would be impossible with these relatively large servos, so tunnels are formed in the bones at their origin points instead, and the tendons are laced through these to preserve the biological orientation of their muscles. This way, every muscle can be realized with extrinsic servos, as wrist movements – which would affect tendon path lengths – are not implemented yet, and they can be corrected in the control algorithm in the future. Similarly, this method results in the transformation of the mechanical dependency between the FDP and LUM into the digital space, enabling coordination between their activities to be managed by the control software.

As shown on Fig. 2/(b), 1 mm thick silicone of ShoreA60 hardness is used to mimic the elastic pulley system, screwed to the bone shafts and the MCP volar plate with M1 screws in multiple points. PTFE tube pieces are attached to them to serve as tendon guides and prevent the strings from tearing into the silicone. Flexor tendons are realized as 0.5 mm diameter strands of highly resistant shoe sewing thread, which are connected to sunken screws in their anatomical insertion points, after being laced through the pulley system. The flexor and extensor retinacula are fabricated and fixed to the carpal bones similarly to the pulley system, following their biological, multi-tunneled structure, with separate tube pieces serving as sheaths for each tendon. The extensor network was carefully designed to mimic the anatomically determined geometry and to retain the location of the biological insertion points, as shown in Fig. 3/(b), with the previously described important functional characteristics in mind. It is laser-cut as a single piece from ShoreA80 silicone of 1 mm thickness to contain every previously described slip, band and ligamentous extension, wrapped around the finger, attached to the bones with M1 screws, and connected to the servos by thread. The two-fold structure of the extensor hood was mimicked by a continuous piece of sheet at the thicker, and a ribbed area at the thinner section. The deep expansion and the first insertions of the INT to the PP base – which are the only "protruding" areas from the flat, laser-cut extensor network – are realized by the thread continuing from the primary thread insertion at the silicone tendon endings, connecting the tendons directly to the screws sunk into the bone surfaces. This design allows for the examination of every previously described behavior: the posterior PIP joint recess was included to enable unobstructed sliding of the lateral bands up to and substituting the point limited by the TRL, and the thickness and angle between the different bands facilitate their individual tightening and relaxation in response to intrinsic and extrinsic muscle activity, promoting their varying effects following the changes in joint angles, while retaining passive movement range limitations that are similar to the human hand. The silicone network is solely responsible for transmitting torques for all finger movements concerning the extensor system, with no further string-based connection distally to the EDC and INT insertions at the PP base, therefore it honestly translates the biological functionality of these features.

### D. Biomimetic Thumb

The opposable thumb, being essential for the formation of most grips, plays a critical and unique part in hand functionality. It has a deterministic role in almost all precise manipulations, and thus it requires considerably more mechanical flexibility and control over its joint positions than any of the other digits. Unlike the long fingers, it has only one IP joint, and its CMC joint is strongly relied upon during almost all of its possible movements.

The TMC joint is composed of two saddle-shaped articular surfaces, the inner ridges of which are transposed asymmetrically. This displacement results in the line of movement along the axis lying perpendicularly to the concave plane of the trapezium being slightly oblique medially, which causes a small degree of axial rotation of the MC bone during thumb opposition. This results in the fingertip trajectory following a curve upon flexion of the TMC, automatically orbiting the long axis of the index finger, and enabling frontal contact between the opposed fingertip surfaces. Its ligamentous structure enables considerable play, which allows the MC bone to take almost any position in space that is not limited by the contacting surfaces [3], [22], and ensures that the bones stick to these boundaries even under passive strain. The mechanical consequences of this structure are of the primary reasons behind the importance of the thumb, their limiting factors being just as important as the laxness and freedom of movement at the joint. They allow for a different degree of displacement and rotation in each direction, further constrained by the irregular surface geometry, meaning that the axes of movement are in fact, not fixed, which is why truly human-like movements would be difficult to achieve by a simple ball joint, or a symmetric saddle joint with generalized ligaments.

The MCP joint bears strong structural similarities to the same joints of the long fingers, both regarding articular surface geometry and ligamentous structure, although the displacements of the thumb's PP are much less constrained by the collateral ligaments. In intermediate positions of flexion, both the PCL and ACL are lax, allowing for slight active and significant passive axial rotation (pronation and supination), and simultaneously adduction or abduction. Another notable difference is the addition of two small sesamoid bones [23] to the anterior side of the volar plate. These serve as focus points for thenar muscle activity, distributing torques through thin ligaments to different regions of the MCP joint to assist its flexion and control displacements of the PP. These features, along with the previously mentioned conical axis of the IP joint, serve to enhance opposition and our fingers' ability to conform around irregular objects, even by altering the shape and orientation of the palmar tunnel.

For taking advantage of the highly capable joint structure, the thumb utilizes an abundance of extrinsic and intrinsic muscles, as shown in Fig. 4/(a). It has a single extrinsic flexor, the flexor pollicis longus (FPL), which arrives from the carpal tunnel, and traverses under its own pulley system on the MC and PP shafts before reaching its insertion at the base of the DP. It synergizes with the three extrinsic extensors to coordinate the major movements of the thumb, acting as a strong flexor at every joint simultaneously, with a single extensor antagonizing it at each (the abductor pollicis longus (APL), extensor pollicis brevis (EPB) and extensor pollicis longus (EPL), the extensors of the TMC, MCP and IP joints respectively), allowing for their independent flexion-extension, and eliminating the need for an elaborate extensor network.

The intrinsic muscles of the thumb, forming the bulk of the thenar eminence, can be divided into a lateral and medial group in accordance with the location of their insertion points. These groups are partial antagonists of each other, and their actions synergize to produce and precisely control the opposition of the thumb. The medial sesamoid muscles (adductor pollicis (AP),

TABLE II
FUNCTIONAL ANATOMY OF THUMB MUSCLES

| Muscle | Function | Critical modeling details |
|---|---|---|
| Flexor pollicis longus | Flexion of all joints simultaneously, beginning with the more distal ones | Arrives from the carpal tunnel, flexes thumb in ulnar direction |
| Thumb extensors (APL, EPB, EPL) | Extension and abduction of every joint individually (TMC, MCP, IP respectively) | Distal end of EPL is coadunate with the extensor expansion, two separate tunnels under ER |
| Adductor pollicis (medial) | Adduction of the MC, lateral rotation of PP, turning it increasingly towards the last digits | Medial sesamoid muscle, affects sesamoid and PP base, two heads: coursing from capitate and 2nd/3rd MC shaft |
| First anterior interosseous (medial) | Adduction of the MC, extension of the IP joint by pulling the extensor expansion | Affects PP base and extensor expansion, superficial layer, arising from 1st/2nd MC shaft |
| Abductor pollicis brevis (lateral) | Anteromedial MC movements, MCP flexion, PP lateral tilt and medial rotation, IP extension via the expansion | Superficial layer, anteromedial to the MC bone, affects PP and the expansion, source on FR and scaphoid |
| Flexor pollicis brevis (lateral) | Adduction to the last digits, flexion of MCP | Deep layer, lateral sesamoid muscle, affects sesamoid and PP base, two heads: coursing from trapezium and trapezoid |
| Opponens pollicis (lateral) | Anteposition and pronation of the MC, adduction to the other digits, primary effector of opposition | Intermediate layer, posterolateral course, pulls lateral border of MC shaft, originates from FR and trapezium |

Table II lists all thumb muscles, along with their notable individual functions and the important modeling details concerning their location and geometry.

and the first anterior interosseous (AI-1)) coordinate opposition if we aim to bring the thumb's fingertip into contact with the index, primarily by the adduction of the MC and rotation of the PP. Meanwhile, the lateral group (flexor pollicis brevis (FPB), abductor pollicis brevis (APB), and opponens pollicis (OP)) is mostly active when the degree of opposition is greater, moving the MC under the palm and medially rotating and tilting the PP. Although the thumb lacks the complex extensor network of the long fingers, both groups have designated muscles that form an extensor expansion that wraps around the PP and merges with the EPL tendon in the middle, enabling the thenar groups to extend the IP joint, while also promoting MCP flexion. It can be concluded that opposition is effectively formed by the synergistic activity of these muscles: the extrinsics exert most of the thumb's gripping power, while the intrinsics contribute to precisely positioning the MC bone by turning its anterior surface towards the other digits, while displacing and rotating the PP to face the fingertips. Their exact biomechanical roles are further detailed in Table II.

The complexity and importance of thumb physiology [3] indicates that if we aim to accurately model its dexterous behavior, we need to mimic its detailed anatomical structure, to provide the same freedom of movement in the joints, and the means to actuate it by correctly realizing the orientation of the multi-layered thenar muscular system. As shown in Fig. 4/(b), all of the TMC ligaments (IML, OPML, OAML, and SALL [3]) were modeled in their anatomical position by 1 mm thick silicone strips of ShoreA60 hardness, while the sesamoids were realized as tiny 3D printed toruses, tethered to the silicone MCP ligaments and the tendons of the sesamoid muscles. The elastic silicone composition of the small MCP ligaments is designed to affect the entirety of the MCP joint upon the contraction of these muscles, enabling us to accurately model the individual activity of each. The extrinsic muscle strands have their own separate tunnels in the flexor and extensor retinacula, with their more distal pathways following the biological courses designated by the 3D printed carpal bone complex. The extensor expansion of the thumb is laser-cut from ShoreA80 silicone, with its lateral bands connected to their designated intrinsic muscles.

Intrinsic muscle tendons are also realized by strands of 0.5 mm thread, which are laced through tendon sheaths formed in the location of their origin points (as indicated in [24]) either directly inside the carpal bones, or by the addition of small, 3D printed tubes to the bone surfaces and the silicone flexor retinaculum, positioned to mimic the layered thenar muscle complex. Some of these muscles are two-headed, or originate from a wider area (e.g. the tubercle of the trapezium) on the carpal bones, which greatly affects the direction of their exerted forces. To account for this, the general force directions of the wider bellies were averaged to pinpoint the location of the string guides needed to preserve their force lines, or individual heads were realized as independent strands coursing through guides corresponding to their origin points, which continue below the level of the wrist to be actuated by separate servos.

Generally, the muscles of the thumb are also actuated by XL-320s, with the exception of the FPL, EPL and the distal head of the AP, which are using MX-12Ws for their increased resolution, and the APL and EPB, which are substituted with AX-12A servos to benefit from their higher torque rating upon antagonizing the FPL and the thenar muscles.

Our design aims to optimize the fabrication process by eliminating the need for complex manufacturing procedures. Rapid prototyping techniques like 3D printing and laser-cutting can be used to produce all components: Printing takes approximately 8 hours for the hand itself and 30 hours for the forearm compartment, and laser cutting takes around 15 minutes. The assembly workflow is simplified by solely relying on M1 and M2 screws to join the pieces together, thus the complete process can be finished by a single person within three days, if so required. Maintenance and repair is also straightforward: the complete extensor network can be cut from a single piece of silicone and fixed at 5 points by M1 screws, and every soft component (ligaments, pulleys) is easily replaceable, as they are fabricated independently and entirely out of silicone, and strings serve only as direct interconnections between the silicone slips and the motors, facilitating robustness and durability.

## III. VALIDATION AND DISCUSSION

In this section, we present the results of the experiments performed to validate the functional capabilities of our prototype. First, we compare the extrinsic tendon excursions related to flexion-extension movements with the biomechanical models of Landsmeer. Then we analyze the synergistic muscle activity used to attain certain finger postures, evaluate the index and thumb fingertip trajectories during wide movements covering their workspaces, and perform the Kapandji test for functional validation. Finally, we demonstrate our hand's prehensile capabilities by realizing the GRASP taxonomy.

### A. Biomechanical evaluation

Landsmeer's models I, II and III [25] are often used in robotic hand research to validate the biomechanical relations between tendon excursions, muscle efficiency and joint angles [6], [15], [26]. We use them to verify that our hand operates with a similar tendon excursion range as described in the biomechanical model.

1) Model I (1) is used for describing situations where the tendon follows the curvature of the articular surface, so it is fitting for the validation of the EDC's effects on the MCP joint:

$$E = r \times \theta \quad (1)$$

where $E$ is the tendon excursion, $r$ is the radius of curvature, and $\theta$ is the joint angle.

2) Model II (2) has the tendon running through a loop, which is freely movable around the axis. The two parts of the tendon run parallel to the long axis of the bone, with the sling being in a position along the bisection of the joint angle. In the validation of flexion-extension by the FDP, FDS and EDC, this model was not used, as the other two models were deemed more suitable.

$$E = 2r \sin\left(\frac{\theta}{2}\right) \quad (2)$$

which works with the same parameters as (1).

3) In Model III (3), the tendon runs in a sheath (pulley), which constrains its path close to the shaft of the bone, but allows it to curve below the area of the joint angle. This fits the the FDP and FDS, since both tendons are coursing through the palmar pulleys of the phalanges. However, as they affect

TABLE III
BIOMECHANICAL COMPARISONS

| Muscles | Parameters | Values |
|---|---|---|
| FDP | MCP $\theta$ (°) | 79.4 |
|  | PIP $\theta$ (°) | 97.5 |
|  | DIP $\theta$ (°) | 74.7 |
|  | calc. $E$ (mm) | 31.4 |
|  | meas. $E$ (mm) | 32.7 |
| FDS | MCP $\theta$ (°) | 84.5 |
|  | PIP $\theta$ (°) | 91.1 |
|  | calc. $E$ (mm) | 23.8 |
|  | meas. $E$ (mm) | 26.1 |
| EDC | MCP $\theta$ (°) | 109.9 |
|  | calc. $E$ (mm) | 14.7 |
|  | meas. $E$ (mm) | 14.3 |

Table III contains the appropriate, calculated and measured tendon excursions during full flexion and extension. Landsmeer's model I was used for the EDC, and model III was used for the FDP and FDS. Deviance between calculated and measured EDC values is notably smaller, since this system has less variables influenced by elasticity and organic bone shapes.

multiple joints simultaneously, the equation was first calculated at each affected joint of a tendon, then the results were summarized to calculate the total tendon excursions.

$$E = 2y + \theta \times d - \theta \times \frac{y}{tan\left(\frac{\theta}{2}\right)} \quad (3)$$

where *y* is the distance of the end of the tendon sheath from the joint center, and *d* is the constraining distance of the sheath from the bone shaft.

We use these models to predict the tendon excursions required to perform movements of flexion and extension with the index finger, and compare the resulting values to actual measurements taken from actuating of our hand model to reach the same joint angles. Our results are listed in Table III. It can be deduced that our measurements are in close proximity of the calculated values, biomechanically validating our approach. The differences can be explained with the fact that biological relations between the elastic human soft tissues are inherently more optimized than the artificial structure used for their approximation, and that the models approximate the organic bone shapes with simplified geometry.

*B. Analysis of long finger movements*

We aim to further validate our approach by describing a few examples commonly used by robotic hand designers for validation purposes [6], [11], [27]. We explain how the general movement components are coordinated primarily by the extrinsic muscles, and how they rely on their synergistic activity with the INT and LUM to precisely and independently control joint positions. At the same time, we demonstrate the functional viability of our design by performing these movements on the index finger of our biomechatronic hand, by individually controlling the servos assigned to each referred muscle, as described in this subsection.

The extrinsic flexors and extensors act together to balance flexion-extension movements. Their rate of contraction determines the degree of flexion in the joints, relying on the extensor network to distribute the generated torques. The EDC's primary activity extends the MCP via its deep slip, and can also partially extend the relaxed IP joints via its medial and lateral bands, which cannot fully tense up before the EDC completes the extension of the MCP. However, this functionality is altered when the IP joints are flexed. The deep expansion becomes slackened as the EDC tendon is pulled forward, so the EDC loses its direct effect on the MCP joint. Flexing the PIP joint causes the lateral bands to slide down to the sides around the PP head, eventually passing the axis of rotation, as the joint is flexed continuously, essentially partially transmitting EDC torques for PIP joint flexion enhancement, balancing out the pulling forces on the central slip. The EDC can still keep the MCP extended by antagonizing the effects of the FDP, but by acting on the PIP joint, since the deep slip is slackened. This synergistic mechanism causes the formation of the claw finger, shown in Fig. 5/(a), enabling the FDP to flex the PIP and DIP joints even while the EDC is contracted to keep the MCP in extension, allowing for the independent flexion-extension of the MCP and the IP joints, without specific extrinsic muscles being dedicated exquisitely to this purpose. If the EDC is relaxed, the FDP, supported by the FDS, acts on the MCP joint as well, resulting in full flexion, shown in Fig. 5/(b). In both cases, the sliding mechanism of the lateral bands helps to coordinate simultaneous IP joint flexion, which is in turn supported by the ORL during extension.

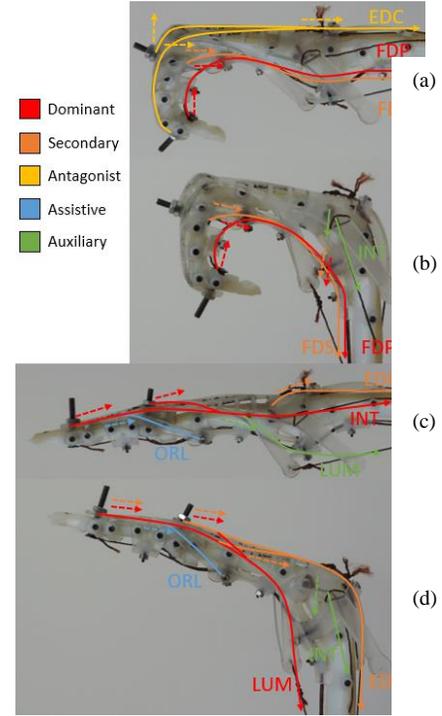

Fig. 5. Examples for synergistic movement of the index finger, as performed by our artificial finger, highlighting the most prominent muscle activities and the acting direction of each participating tendinous band (full lines). The directions of the exerted forces at the tendon contact points are denoted with dashed lines. (a) Claw finger with the EDC antagonizing the long flexors at the MCP joint. (b) Full flexion by strong activity of the FDP and FDS, and assistive action of the INT at the MCP joint. (c) Full extension with the INT pulling on the lateral bands, and assisting the EDC in the extension of the IP joints. (d) Beak position, formed by IP extension during MCP flexion, showing the effects of LUM activity. The EDC acts as an extensor on the IP joints, enhancing the effects of the LUM, while the INT strengthen MCP flexion.

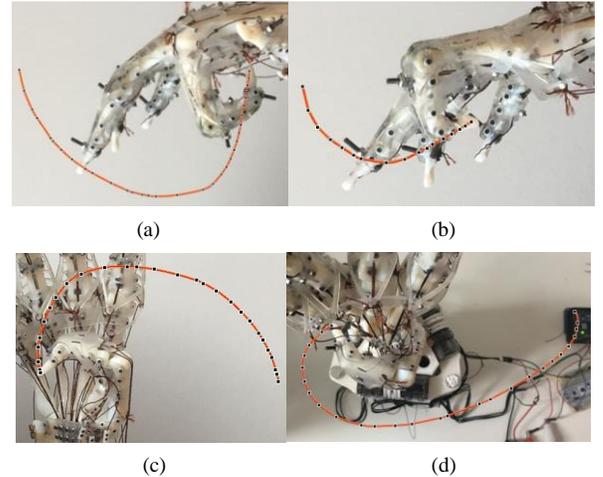

Fig. 6. Fingertip trajectories of the ACB-Hand. (a) Full flexion of the index finger, (b) claw index finger, (c) full flexion of the thumb parallel to the palmar plane, (d) full distal thumb opposition to the 5[th] MC bone head.

The intrinsic muscles, beside their leading role in adduction and abduction, synergize with the EDC to extend the IP joints or enhance MCP flexion, depending on the current angle of the latter. If the MCP joint is extended by the EDC, the extensor hood is carried past proximally over the MCP joint. The proximal tendinous slips of the INT lie almost parallel with the lateral EDC bands, so they can extend the IP joints upon contraction by pulling on the lateral bands and medial slips, resulting in full finger extension, as shown in Fig. 5/(c). If, however, the MCP joint is flexed and the EDC is relaxed, the bulk of the extensor hood moves distally over the MCP joint, to lie at an almost perpendicular angle to the interosseous muscles. In this position, the INT course under the MCP flexion-extension axis, so during contraction they press the hood tightly to the surface of the proximal half of the PP, significantly enhancing flexion strength at the MCP. The lateral bands are slackened as a consequence, so the INT lose their extending effect on the IP joints. The deep expansion of the EDC also relaxes, and the slackening of the proximal segments of the INT lateral bands allow the EDC to tighten up their distal segments. Thus, the EDC contributes to the extension of the IP joints instead.

This posture, shown in Fig. 5/(d), which is used when forming a beak-like shape with our hand, is further supported by the activity of the LUM. They lie anterior to the DTML, so their contact point is at an increased angle with the lateral bands, which means that they can flex the MCP even if it is hyperextended. Additionally, they are not tethered by the extensor hood, so they can assist the EDC in tightening up the lateral bands and extending the IP joints even when the interosseous muscles hold the extensor hood down to flex the MCP. They also promote the extension of the IP joints due to them originating from the FDP tendon, so when they contract, the distal part of the tendon becomes slackened. This actively transmits the torque of the FDP muscle to the lateral bands, preventing it from flexing the IP joints, enhancing their extending effect on them instead. Since their activity does not require FDP contraction, but their effects are enhanced by it, they promote independent MCP flexion and IP extension, without severely affecting or being affected by the joint angles in the other fingers. This makes them invaluable during dexterous manipulation tasks, where a high degree of independent finger control is required, especially in cases with significant MCP flexion, but only moderate IP flexion.

These examples serve to validate our design by showing that the robotic hand is able to adequately produce even more complex, synergistic movements, just as we would expect from the specific role of each controlled muscle in human functional anatomy. Therefore, we hope that the model can be useful as a platform in medical research, and for biomechanical analysis.

*C. Examination of fingertip trajectories*

As we previously described, the flexion-extension of the fingers results from the synergistic activity of the FDP, EDC and FDS, with the addition of the INT, and optionally the LUM to enhance flexion or achieve full extension. Furthermore, the trajectory followed by the fingertip depends on the activation rate of the extrinsic muscles. We examine the two most extreme variations of flexion in the index finger: full flexion (Fig. 6/(a)) and the claw finger (Fig. 6/(b)). To identify the required flexor muscle activity to produce these movements, measurements were taken on a human operator using a SonoScape A6V ultrasound sensor. It was observed that the claw finger is produced by the FDP, which is antagonized by the EDC at the MCP joint, with an inactive FDS. In full flexion, however, both flexors are active, with the FDS strengthening flexion at the MCP. We control the three respective actuators accordingly, showing that in both cases, simultaneous flexion is observable in the PIP and DIP joints, either due to the coactivity of the FDP and FDS, or the sliding mechanism of the lateral bands and the laxness of the medial band due to the taut deep slip.

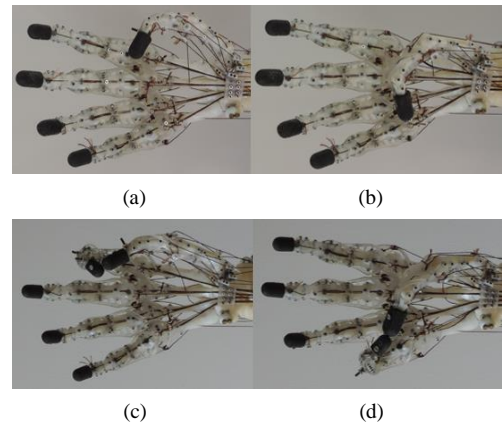

Fig. 7. Extreme positions of the Kapandji test, as performed on the ACB-Hand. Touching the (a) pad of the $2^{nd}$ MCP, (b) pad of the $5^{th}$ MCP, (c) tip of the index, (d) tip of the little finger with the fingertip of the thumb.

In the case of the thumb, there is no elaborate extensor network to be examined thoroughly. During object manipulation and grasping, the extrinsic muscles are responsible for broadly coordinating the movements of flexion-extension and abduction, while the intrinsics precisely control adduction, axial rotation and lateral or medial displacement at the MCP joint, as summarized previously in Table II. However, we can distinguish between two generally different thumb movements with regard to grasping: flexion (Fig. 6/(c)) and opposition (Fig. 6/(d)). Flexion is performed primarily by the FPL, occurring upon the lack of antagonist activity by the extrinsic extensors at the joints, with the medial intrinsic muscle group assisting in the adduction of the MC. As a result of this, the plane of the flexion trajectory lies almost parallel to the palm. In full (distal) opposition, the effects of the FPL are less remarkable, contributing mainly to the slight flexion of the IP and the MCP, as first the MC is opposed and abducted by the OP, then the MCP is flexed and pronated by the other members of the lateral muscle group. This results in the thumb's fingertip effectively circulating around the index finger. Our model succeeds at accurately producing flexion by actuating the distal and proximal heads of the AP separately to assist the FPL. The experiment highlights the precise synergy needed during wide opposition, where the EPB and the APB first enhance the abducting effects of the OP, but then need to synergistically relax as the FPB, FPL, and in certain orientations even the AI-1 contract to move the fingertip towards the last MC head.

Following this analysis, we individually control the servos during each experimental movement with a prepared command sequence, and use a mounted camera to capture them on photo bursts of 10 images per second. Then we analyze the frames to pinpoint the location of the fingertip, and examine the resulting 2D trajectories (Fig. 6), from which we conclude that the artificial fingers of the ACB-Hand are capable of closely following the biomechanically assumed trajectories in these simple examples, using similar muscle activity as humans during biological movements, validating our anatomically accurate realization for general finger motion.

Fig. 8. The realization of the 33 grasps described in the GRASP taxonomy [29] with the ACB-Hand.

### D. Performing the Kapandji test

The purpose the Kapandji test [28] is to medically assess thumb opposition, by evaluating whether certain areas of the hand are reachable by the fingertip of the thumb. We performed the stages of the test for the palmar pads of each MCP joint, and the fingertip of each digit. As expected based on the anatomically correct thumb implementation, we were able to successfully execute all attempted stages. When the thumb's fingertip is brought into contact with different areas of the index finger, varying degrees of flexion are required without significant opposition, whereas when reaching the little finger, the thumb is more heavily opposed by the activity of the OP. These extreme positions are shown in Fig. 7.

### E. Grasping demonstration

To evaluate the overall functional performance of our hand model, we attempt to perform the definitive grasping examples classified in the GRASP taxonomy [29], using our manual control system, where we can assign any individual actuator or group to a button-based control board. As shown in Fig. 8, during most power grasps, our prototype was able to maintain a firm grip around the object without any additional assistance. To more efficiently execute precision grips where the point of contact is small or the gripping forces are concentrated at the fingertips, the application of soft, 3D printed (Tango Plus FLX, PolyJet by Varinex Zrt.) fingertips was required to provide softness and additional friction. Using these, the ACB-Hand could reliably perform even during grasps which would rely on the additional palmar volume provided by muscle mass and the extra friction on the surface of the skin (tripod variation, palmar and adduction grip), and thus successfully form a firm and stable grasp for all 33 cases in the taxonomy.

## IV. CONCLUSION AND FUTURE WORK

With the development of the ACB-Hand, we have prototyped a competent, anatomically correct, biomechatronic interpretation of the human hand, which includes all functionally relevant bones, muscles, tendons and ligaments for finger actuation, and shown how the dexterous capabilities of a robotic model can benefit from the inclusion of the lumbrical and interosseous muscles, alongside the extensor network and the thenar muscles. We have shown that our proposed design satisfies the biomechanical requirements and dependencies behind the behavior of the hand: we have described the four primary finger movements that require synergistic-antagonistic coordination between the extrinsic and intrinsic muscles (full flexion, full extension, claw finger and beak position), and shown that our definitive anatomical approach enables moving each muscle in accordance with their assumed biological role, resulting in functionally correct, synergistic movement execution, while closely following the human fingertip trajectories. We applied the commonly used biomechanical models of Landsmeer, and validated our design by showing that our measured tendon excursions are closely following the calculations based on our design's parameters. We have also validated the hand's ability to correctly produce human-like thumb opposition by performing the Kapandji test, and shown that the taxonomical requirements of efficient and versatile grasping are satisfied by successfully performing the 33 examples specified in the GRASP taxonomy. We hope that this device can help us better understand the direct biomechanical influence of human anatomical structure on high-level functionality, to aid in the development of advanced manipulators and hand prosthetics.

In the future, we plan to explore new ideas to reduce the overall size of the actuation system, and profit upon this upgrade to implement a competent wrist mechanism. At the same time, we aim to apply a suitable artificial layer of skin, to promote the formation of stable palmar grips. Aside from the mechanical considerations, we will focus on studying the required muscle activity for more complex movements towards developing a control interface for teleoperation. We will also collaborate with neuroscientists and neuroengineers to explore possibilities for the design's utility in the development of neural interfaces for precise, independent muscular control, and to further highlight the advantages of anatomical feature inclusion in commercial prosthetic design, which could hopefully be a valuable asset in the quest for truly restoring hand dexterity.

ACKNOWLEDGMENT

The authors would like to thank the colleagues of the Robotics Laboratory at PPCU-FIT for their help during the various stages of the workflow, Varinex Zrt. for their assistance in the production of several 3D printed components, and Dr. Béla Novoth leading hand surgeon for his assistance and special insight into the field of human hand anatomy.